\documentclass{article}
\usepackage[submission]{colm2026_conference}

\usepackage{hyperref}
\usepackage{url}
\usepackage{booktabs}
\usepackage{lineno}
\usepackage{color}
\usepackage{tcolorbox}
\usepackage{caption}
\tcbuselibrary{skins}

\usepackage{amsmath,amsfonts,bm}

\def\eqref#1{equation~\ref{#1}}

\def\1{\bm{1}}

\DeclareMathAlphabet{\mathsfit}{\encodingdefault}{\sfdefault}{m}{sl}
\SetMathAlphabet{\mathsfit}{bold}{\encodingdefault}{\sfdefault}{bx}{n}

\usepackage{xcolor}
\usepackage{graphicx}
\usepackage{tabularx}
\usepackage{amsmath}
\usepackage[capitalize]{cleveref}
\usepackage{mathtools}
\usepackage{multirow}
\usepackage{setspace}
\hypersetup{colorlinks,linkcolor={blue},citecolor={blue},urlcolor={blue}}

\newif\ifshowcomments
\showcommentsfalse

\definecolor{absgray}{RGB}{242,243,245}
\definecolor{metablue}{RGB}{0,102,204}

\newcommand{\customabstractpage}{
\begin{tcolorbox}[
    enhanced,
    colback=absgray,
    colframe=absgray,
    boxrule=0pt,
    arc=8pt,
    left=3mm,
    right=3mm,
    top=3mm,
    bottom=3mm
]

{\Large\bfseries
Alignment Tuning for Large Language Models: \\A Data-Centric Lens on Alignment Data Pipelines
\par}

\vspace{3mm}

Hwanjun Song\par

\vspace{1mm}

KAIST\par

\vspace{4mm}

\noindent
Much of the alignment tuning literature is organized around optimization objectives, while the construction of alignment data is often treated implicitly. In this survey, we adopt a data centric perspective and reframe alignment tuning as a pipeline design problem. We decompose alignment data construction into three interacting stages, response synthesis, preference evaluation, and preference instantiation, and use this framework to organize existing alignment methods into a unified taxonomy. Through this lens, we identify recurring design trade-offs and failure modes observed across prior alignment methods, and distill a set of high level principles that clarify how pipeline design choices influence the resulting optimization signal. Finally, we outline open challenges for alignment data pipelines, including prompt-level alignment, agentic settings, and alignment under evolving objectives.

\vspace{4mm}

\noindent
\begin{minipage}[t]{0.70\textwidth}
{\small
\textbf{Date:} April 6, 2026 \par
\textbf{Correspondence:} Hwanjun Song at {\color{metablue}\href{mailto:songhwanjun@kaist.ac.kr}{songhwanjun@kaist.ac.kr}} \par
}
\textbf{Publication:} Will appear at the Findings of ACL`26.
\end{minipage}
\hfill
\begin{minipage}[t]{0.27\textwidth}
\vspace*{-0.5cm}
\raggedleft
\includegraphics[width=1.0\linewidth]{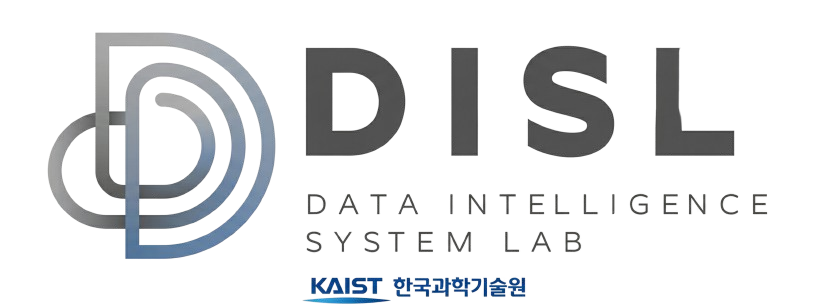}
\end{minipage}

\end{tcolorbox}
}

\begin{document}

\thispagestyle{empty}

\vspace*{-1.3cm}
\customabstractpage

\section{Introduction}

The progress of large language models (LLMs) has been driven by scaling laws, enabled by increased model parameters \citep{kaplan2020scaling}, architectural innovations \citep{fedus2022switch, wu2024multi}, and advances in optimization \citep{yu2025dapo}.
As the marginal returns of scaling plateau, performance gains have shifted toward data-centric factors, with data quality emerging as a key driver \citep{chung2024scaling, zhuang2025meta, nazar2025design}.
Despite this shift, most prior work views data quality through \emph{static corpora}, focusing on dataset composition and filtering for pre-training or supervised fine-tuning (SFT) \citep{brown2020language, liu2024coachlm, liuregmix}.


However, this static perspective is insufficient to explain the safety, robustness, and preference adherence of modern LLMs.
These properties are primarily shaped during \emph{alignment tuning}, a post-training phase distinct from pre-training \citep{ji2023ai, rafailov2023direct, bai2025online}. 
Unlike supervision from fixed distributions, alignment data is inherently \emph{dynamic} and \emph{policy-dependent}, generated through repeated interactions among prompts, model outputs, and feedback signals \citep{li2025optimizing, yu2025cot, liu2025spice}. 
As a result, alignment quality is governed less by static data artifacts and more by the mechanisms that iteratively construct and evaluate them.

\begin{figure*}[t!]
\begin{center}
\includegraphics[width=14cm]{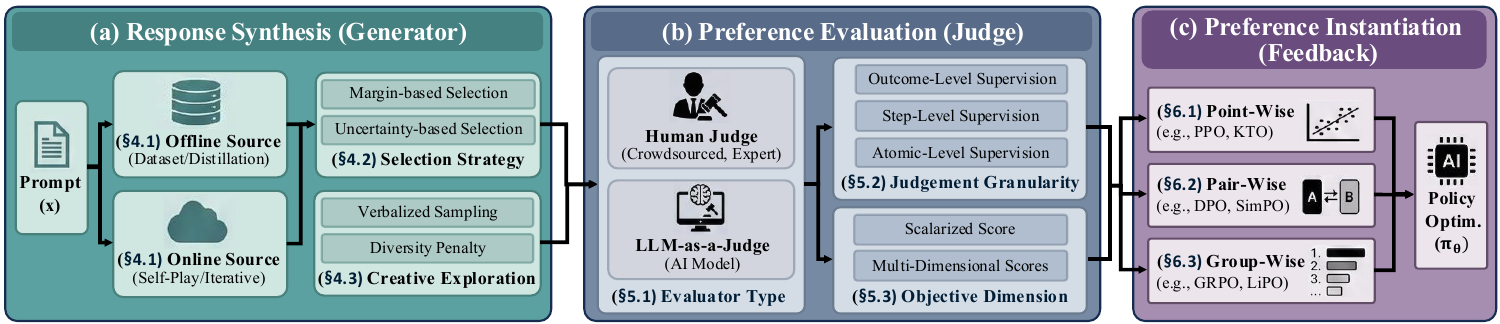}
\end{center}
\vspace*{-0.40cm}
\caption{Overview of the alignment data pipeline, showing how prompts are converted into structured optimization signals through response synthesis, preference evaluation, and preference instantiation for policy optimization.}
\label{fig:overview}
\vspace*{-0.25cm}
\end{figure*}

We therefore conceptualize alignment tuning not as a dataset curation task, but as a \emph{pipeline design} problem. Data quality in alignment tuning depends not only on which samples are retained, but on how candidate behaviors are generated, evaluated, and structured into learning objectives. We introduce a unifying framework with three interacting dimensions: response synthesis, preference evaluation, and preference instantiation. Figure \ref{fig:overview} summarizes this, highlighting how tightly coupled stages jointly construct the optimization signal.


\smallskip \noindent \emph{(1) Response Synthesis:}  This stage defines the behavioral support of alignment by determining how candidate responses are generated. Key design choices include the response source (offline distillation versus online sampling) \citep{rafailov2023direct, zhangself, yu2025diverse}, selection strategies that prioritize informative candidates, and exploration mechanisms that preserve diversity and avoid premature mode collapse \citep{wuself, lanchantin2025diverse}.


\smallskip \noindent \emph{(2) Preference Evaluation:} Given synthesized responses, alignment depends on the fidelity of preference signals. This dimension spans evaluator type, from human annotation to scalable LLM-as-a-Judge frameworks \citep{lee2024rlaif, yu2025diverse}, as well as judgment granularity and objective dimensionality, which determine preference fidelity and the risk of reward hacking or alignment tax under scalarized and coarse supervision \citep{li2025gradient, mukherjee2024multi}.


\noindent \emph{(3) Preference Instantiation:} Finally, preference instantiation determines how evaluative judgments are exposed to optimization.
This includes point-wise rewards \citep{ethayarajh2024kto, yuan2024self}, pair-wise contrasts \citep{rafailov2023direct, meng2024simpo}, and group- or list-wise formulations \citep{ramesh2024group, liu2025lipo}, which differ in how effectively preference structure is translated into policy updates.


\smallskip 
From this perspective, we organize prior work into a unified data-centric taxonomy (Sections \ref{sec:synthesis}--\ref{sec:instanciation}) and distill a set of design principles that characterize recurring trade-offs and cross-stage interactions across data pipeline stages (Section \ref{sec:insight}). 


\smallskip
\textbf{Related Surveys.}  
Existing surveys on data-centric LLM training primarily emphasize static data stages, such as data selection for pre-training and SFT \citep{albalak2024a, wang2023data}, dataset catalogs \citep{liu2025datasets}, or general training paradigms \citep{minaee2024large}, as well as system-level considerations \citep{xu2024data, zhou2025survey}.
In contrast, we focus on alignment tuning as a dynamic, closed-loop pipeline, examining how response synthesis, evaluation, and instantiation jointly shape alignment outcomes.

\smallskip
\textbf{Our Scope.}
Section \ref{sec:algorithm} provides a brief overview of alignment algorithms, but the majority of our analysis centers on the alignment data pipeline. Detailed discussions of optimization algorithms are deferred to existing surveys on alignment techniques \citep{xiao2024comprehensive}, direct preference optimization \citep{liu2025survey}, unified loss design \citep{tang2024generalized}, and fundamental limitations such as reward hacking \citep{casper2023open}.


\section{Alignment Tuning Foundation}

The central challenge in developing LLMs lies in the mismatch between their training objective and human preferences \citep{stiennon2020learning, ouyang2022training, bai2022constitutional}. Standard next-token prediction maximizes data likelihood, which is largely orthogonal to desiderata such as helpfulness, honesty, and safety. As a result, models trained only via pre-training and supervised fine-tuning may exhibit factual errors or harmful behaviors despite high likelihood performance. Alignment tuning addresses this gap by explicitly optimizing models toward human-valued behaviors.

\subsection{Problem Formulation}

Let $x$ denote a prompt sampled from a task distribution $P$, and $y$ be a response generated by a policy $\pi_\theta$. We assume an oracle reward function $r^{*}(x,y)$ reflecting human preferences. Alignment tuning seeks an optimal policy $\pi^*$ that maximizes expected reward while remaining close to a reference policy $\pi_{\text{ref}}$, preventing reward hacking and uncontrolled drift \citep{ji2023ai, yehposition}:
\begin{equation}
\small
\max_{\pi_\theta} \mathbb{E}_{x,y} \left[ r^*(x, y) - \beta \log \frac{\pi_\theta(y|x)}{\pi_{\text{ref}}(y|x)} \right],
\label{eq:alignment_tuning}
\end{equation}
where $\beta$ is the regularization coefficient controlling the deviation from the reference model.

\subsection{Optimization Algorithms}
\label{sec:algorithm}

Various approaches have been proposed to solve Eq.~(\ref{eq:alignment_tuning}). We review three foundational methods that represent the evolution of alignment tuning. See Appendix \ref{sec:algo_detail} for details and additional algorithms.


\emph{Proximal Policy Optimization (PPO)} is the method for explicit reward maximization \citep{schulman2017proximal}. It operates in a two-stage process. First, a reward model $r_\phi(x,y)$ is trained to approximate oracle preferences $r^*(x,y)$ via pairwise comparison data. Second, the policy $\pi_\theta$ is optimized using reinforcement learning to maximize the learned reward. PPO employs a clipped surrogate objective to ensure stable updates, restricting the policy update step size to prevent catastrophic forgetting of the reference distribution.


\emph{Direct Preference Optimization (DPO)} is the method that bypasses the explicit reward modeling stage by deriving the optimal policy for the KL-regularized objective in Eq.~\eqref{eq:alignment_tuning} \citep{rafailov2023direct}. This derivation shows that the reward can be implicitly captured through the log-likelihood ratio between the policy and a fixed reference model. As a result, the optimization reduces to a classification-style objective on paired responses $(y_w, y_l)$, directly increasing the likelihood of the preferred output $y_w$ relative to the dispreferred one $y_l$, without requiring a separately trained reward model.


\emph{Group Relative Policy Optimization (GRPO)} shifts from pairwise comparisons to group-wise optimization \citep{shao2024deepseekmath}. Instead of a learned critic, it estimates the baseline for a response by computing the mean reward of a group of outputs sampled from the current policy $\pi_{\theta}$ for the same prompt. The policy model is updated to increase the probability of responses that perform better than the group average. It reduces the variance associated with external reward models and eliminates the need for a separate value network.


\section{Understanding Alignment Tuning from a Data-Centric Perspective}

While optimization algorithms update the policy model $\pi_{\theta}$, they do not by themselves determine the direction or quality of alignment. Instead, from a data-centric perspective, alignment outcomes are governed by the design of the alignment data pipeline, which specifies the space of candidate behaviors, the mechanism by which they are evaluated, and the structure through which preference signals are exposed to optimization.

\subsection{Alignment Data as Optimization Signals}

Alignment tuning relies on preference signals constructed through a data pipeline rather than given a priori. We formalize how this process yields the optimization signals that ultimately drive policy updates and shape the learned behavior.

\smallskip
\textbf{Formalizing the Alignment Data Pipeline.}
Unlike static pre-training corpora, alignment data is dynamically constructed through an iterative pipeline that couples response generation and preference assessment. We formalize the resulting dataset $\mathcal{D}$ as a collection of structured training instances produced by three interacting components:
\begin{equation}
\mathcal{D}
=\Big\{\big(x,\; \mathbf{y},\; \mathbf{s}\big)
\;\Big|\;x \sim P(x),~\mathbf{y} \sim \mathcal{S}(\mathbf{y}\mid x),~\mathbf{s} \sim \mathcal{E}(\mathbf{s}\mid x,\mathbf{y})
\Big\}.
\label{eq:pipeline}
\end{equation}
Here, $x$ is a prompt sampled from the task distribution $P$; $\mathbf{y}=\{y_1,\ldots,y_k\}$ is a set of candidate responses generated by a {response synthesis} strategy $\mathcal{S}$, which defines the behavioral support available for alignment; and $\mathbf{s}$ represents preference signals assigned by an {evaluator} $\mathcal{E}$, which are subsequently structured through {preference instantiation} to form training signals such as scalar scores,  pairwise preferences, or rankings over $\mathbf{y}$.

\smallskip
\textbf{Optimization as Margin Alignment.}
Alignment algorithms optimize the policy $\pi_\theta$ so that its implicit preferences match the explicit signals encoded in the alignment dataset $\mathcal{D}$. 
Across PPO \citep{schulman2017proximal}, DPO \citep{rafailov2023direct}, and GRPO \citep{shao2024deepseekmath}, this process can be viewed as \emph{margin alignment}, where optimization aligns policy-induced preference margins with observed preference signals.

Given a prompt $x$ with candidate responses $\mathbf{y}$ and preference signals $\mathbf{s}$, alignment tuning aims to adjust the policy $\pi_\theta$ so that its implicit preferences are consistent with the structure and magnitude of the preference information encoded in $\mathbf{s}$, aligning model behavior with the supervision induced by the dataset $\mathcal{D}$. This objective can be abstractly written as:
\begin{equation}
\small
\max_{\theta}\; \mathbb{E}_{\mathcal{D}}
\big[ f\big(M_\theta(x,\mathbf{y},\mathbf{s})\big) \big],
\label{eq:matching}
\end{equation}
where $M_\theta(x,\mathbf{y},\mathbf{s})$ denotes an alignment measure that quantifies how well the policy-induced implicit preferences align with the preference signals $\mathbf{s}$. The function $f$ transforms this alignment measure into an optimizable objective in a way that preserves the relative preference ordering.

\smallskip
\textbf{Pipeline Defines Optimization Signal.}
Eq.~(\ref{eq:pipeline}) and Eq.~(\ref{eq:matching}) together show that alignment outcomes are determined not only by the optimization objective, but by how the alignment data pipeline constructs its inputs. While Eq.~(\ref{eq:matching}) frames alignment as maximizing a margin-based objective $M_\theta(x,\mathbf{y},\mathbf{s})$, Eq.~(\ref{eq:pipeline}) specifies the response candidates $\mathbf{y}$, preference signals $\mathbf{s}$, and relational structure on which this margin is defined. As a result, the pipeline does not merely provide data, but actively shapes the space, scale, and reliability of the preference margins available to optimization, thereby influencing the nature of the learned policy.

\subsection{Overview of Alignment Data Pipeline}

As the optimization signal is determined by how alignment data is constructed, we briefly summarize the structure of the alignment data pipeline.

The alignment data pipeline begins with \emph{response synthesis} (Figure \ref{fig:overview}(a)), which defines the behavioral scope of alignment by generating candidate responses. The pipeline then proceeds to \emph{preference evaluation} (Figure \ref{fig:overview}(b)), which assigns supervisory signals to approximate latent preferences. Finally, \emph{preference instantiation} (Figure \ref{fig:overview}(c)) converts evaluated judgments into optimization compatible training signals. Detailed discussions of each stage and their associated design trade-offs are provided in Sections \ref{sec:synthesis}--\ref{sec:insight}. Additionally, we categorizes all the methods we investigated across the three aspects in Table \ref{tab:categorization}.

\section{Response Synthesis Stage} 
\label{sec:synthesis}

This stage defines the exploration space of candidate responses through the synthesis strategy $\mathcal{S}$. Since alignment operates only on sampled responses $y \sim \mathcal{S}(y \mid x)$, it constrains which behaviors enter alignment. It involves three design considerations: \emph{(\S\ref{sec:response_source}) response source}, \emph{(\S\ref{sec:selection_strategy}) selection strategy}, and \emph{(\S\ref{sec:creative_exploration}) creative exploration}.


\subsection{Response Source} 
\label{sec:response_source}


The first design choice in response synthesis is the {response source} (\emph{i.e.}, where candidate responses are drawn from). This choice determines the distributional relationship between the data policy $\pi_{\text{data}}$ and the learned policy $\pi_\theta$, which in turn affects the effectiveness of alignment signals. Existing approaches have evolved along two philosophies, \emph{offline} and \emph{online}, with techniques developed to mitigate their inherent limitations.

\smallskip
\textbf{{Offline with Policy-Aware Reweighting.}} 
Offline approaches rely on fixed, high-quality responses from stronger models (\emph{e.g.}, proprietary LLMs) or curated datasets. While cost-effective, this setting induces {distributional shift} \citep{xiongiterative, bosehybrid}, as the policy is optimized on responses it is unlikely to generate, resulting in biased value estimation. To mitigate this mismatch, recent methods adopt {policy-aware reweighting}, which re-scales samples by the current policy likelihood (\emph{e.g.}, $\pi_\theta / \pi_{\text{data}}$). They differ primarily in the granularity of reweighting: preference-level weighting in WPO \citep{zhou2024wpo}, reward-difference modulation in RDO \citep{wang2024reward}, weighted reward aggregation across multiple sources in WRPO \citep{yang2024weighted}, and token-level importance weighting in TIS-DPO \citep{liu2024tis}. By down-weighting unlikely responses, policy-aware reweighting reduces off-policy bias and enables reliable offline alignment.

\smallskip
\textbf{Online with Structured Self-Improvement.} 
Online approaches train directly on outputs from the current policy $\pi_\theta$, providing on-policy supervision and reducing distributional mismatch. Yet, they incur high computational cost from online generation and can suffer instability due to low-quality rollouts. Recent work addresses these issues through structured self-improvement. Iterative DPO \citep{xiongiterative} and RS-DPO \citep{khaki2024rs} stabilize training via rejection-based filtering, while self-play methods such as SPIN \citep{chen2024self} and SPPO \citep{wuself} cast alignment as a zero-sum game without external oracles. Further extensions improve efficiency through group-relative feedback in GRPO \citep{shao2024deepseekmath} and active exploration in SELM \citep{zhangself}.


\subsection{Selection Strategy}
\label{sec:selection_strategy}

Another key design choice in response synthesis is the {selection strategy}, which determines how informative instances are selected, either post-hoc (offline) or during response pool construction (online). Existing approaches differ primarily in how informativeness is quantified through \emph{margin}-based or \emph{uncertainty}-based criteria.

\smallskip
\textbf{Margin-Based Selection.} Margin-based selection identifies training instances with high alignment potential, defined as response comparisons for the same prompt (\emph{e.g.}, pairs or small ranked sets) where preferences clearly separate better from worse behaviors. \citet{huanglarger} distinguish explicit margins reflecting preference strength from implicit margins induced by policy likelihood differences, whereas BeeS \citep{deng2025less} uses the two margins independently to filter informative preference instances.  Beyond filtering, MMPO \citep{kim2024margin} incorporates explicit margins directly into the optimization objective via soft targets, allowing the model to account for graded preference strength rather than treating all comparisons equally. Similarly, \citet{yao2025enhancing} improve alignment efficiency by modifying how preference margins are integrated into the loss.


\smallskip
\textbf{Uncertainty-Based Selection.} 
Unlike margin-based selection, uncertainty-based selection targets low-confidence regions of the policy where preference signals are ambiguous or weakly specified. Early methods use predictive entropy or probability dispersion, while recent work adopts structured uncertainty modeling. APL \citep{muldrew2024active} and MAPLE \citep{mahmud2025maple} reduce preference uncertainty via information gain maximization and Bayesian adaptive selection, respectively. In online settings, uncertainty serves as a quality-control signal: IUPO \citep{li2025uncertainty} targets locally ambiguous reasoning steps through token-level uncertainty, while UPO \citep{wang2025self} filters unreliable samples using reward-model uncertainty. More recent methods like UDASA \citep{sun2025uncertainty} further decompose uncertainty into semantic, factual, and value-alignment dimensions with curriculum-style selection.

\subsection{Creative Exploration} 
\label{sec:creative_exploration}
Alignment objectives often overemphasize a narrow set of preferred responses, concentrating probability mass and inducing \emph{mode collapse} that reduces semantic diversity and creativity \citep{kirkunderstanding, murthy2025one}. To address this mode collapse, recent methods intervene directly in response synthesis, beyond informativeness-driven selection. During candidate construction, Verbalized Sampling \citep{zhang2025verbalized} promotes exploration by eliciting multiple plausible responses and their likelihoods, while Spectrum Tuning \citep{sorensen2025spec} trains on diverse valid outputs to control stylistic variation. At the post-hoc selection stage, DivPO \citep{lanchantin2025diverse} preserves high-quality but rare responses via diversity-aware filtering, and \citet{chung2025modifying} reweight the objective to favor unique answers. Finally, CRPO \citep{ismayilzada2025creative} explicitly rewards novelty and surprise alongside utility.



\section{Preference Evaluation Stage} 
\label{sec:evaluation}

\begin{figure}[t!]
\begin{center}
\includegraphics[width=7.71cm]{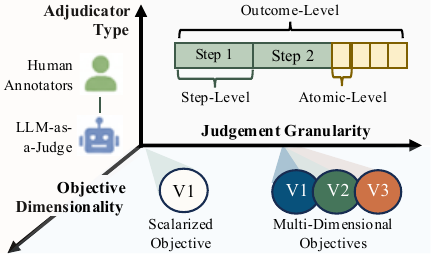}
\end{center}
\vspace*{-0.35cm}
\caption{Preference evaluation axes in alignment data pipelines, illustrating interplay among adjudicator type, judgment granularity, and objective dimensionality.}
\label{fig:axis}
\vspace*{-0.15cm}
\end{figure}

Once response candidates are generated, this stage assigns preference judgments via an adjudicator. While early RLHF relied on human judgments, recent work increasingly adopts automated adjudicators to reduce cost and variability \citep{li2025generation, min2025towards}. Accordingly, prior work can be organized along three design axes in Figure \ref{fig:axis}: \emph{(\S\ref{sec:adjudicator}) adjudicator type}, \emph{(\S\ref{sec:granularity}) evaluation granularity}, and \emph{(\S\ref{sec:dimensionality}) objective dimensionality}.

\vspace*{-0.1cm}
\subsection{Human to AI Adjudicator}
\label{sec:adjudicator}
The first axis is the source of the preference signal. While human evaluation was traditionally considered the {gold standard}, recent findings indicate that inter-rater agreement is often insufficient for complex reasoning or coding tasks, thereby limiting the upper bound of learnable reward signals \citep{kocmi2023large, li2024llms}. 

\smallskip
\textbf{{Use of LLM-as-a-Judge.}} 
The field has shifted toward LLM-as-a-Judge frameworks \citep{li2024llms}, in which LLMs are used as automated evaluators to score or compare candidate responses, improving scalability and reproducibility. These frameworks have been instantiated in diverse way. For example, RLAIF \citep{lee2024rlaif} simply uses off-the-shelf LLMs as automated preference annotators, providing comparison-based feedback for RL, while Constitutional AI \citep{bai2022constitutional} employs a critique-and-revise loop guided by natural language principles to generate preference labels. More recently, Self-Rewarding LMs \citep{yuan2024self} integrate the judge capability directly into the policy model, enabling it to self-assign rewards during iterative alignment training.

\smallskip
\textbf{{Bias Sources and Mitigation.}} 
A critical challenge in LLM-based automated evaluation is intrinsic bias, where the judge's internal priors distort evaluation outcomes \citep{liu2023g, zheng2023judging}. Recent work frames this as a calibration problem, using algorithmic interventions to neutralize specific biases. Specifically, \citet{wataoka2024self} show that LLMs favor outputs from their own model family, exhibiting an “echo chamber” effect, and propose cross-family evaluation using architecturally distinct LLMs as a mitigation strategy. \citet{wang2024large} reveal a pronounced verbosity bias, where LLM judges favor longer responses, and mitigate it through length normalization and constrained evaluation. \citet{ahrabian2025practical} further identify position bias, where models preferentially select the first option, and mitigate it through permutation-based evaluation.

\smallskip
\textbf{{Collective Intelligence.}} 
Despite calibration efforts, single model judgments remain inherently unstable, as they conflate prompt sensitivity and stochastic variation into a single evaluation signal. \citet{yu2025diverse} aggregates judgments from heterogeneous LLM evaluators to average out individual noise. Hierarchical schemes like {Meta-Rater} \citep{zhuang2025meta} further refine this by employing a superior meta-judge to adjudicate conflicts among primary judges. More interactive paradigms instantiate consensus-seeking through multi-agent debate \citep{du2023improving, chen2025multi}, where judges iteratively critique and refine evaluations to yield stable supervision signals. These collective mechanisms improve fidelity to the oracle reward, preventing evaluator noise from being treated as an optimization signal during alignment tuning.

\subsection{Judgment Granularity} 
\label{sec:granularity}
The second axis concerns judgment granularity (\emph{i.e.}, how preference signals are formed within a response). Early methods rely on outcome-level supervision, collapsing response quality into a single global judgment \citep{li2024llms, liu2023g}. While sufficient for short tasks, this is inadequate for long-form and multi-step generation, where errors and merits are unevenly distributed \citep{wang2024math, yan2025atomic}. Recent work thus increases granularity, shifting from outcome-level supervision to localized signals.


\smallskip
\textbf{{Step-Level Supervision.}}
Step-level supervision increases judgment granularity by assigning preference signals along the generation process rather than only at the final outcome \citep{zheng2025survey, li2025generalist}. It is most effective in domains with explicit intermediate structure, such as mathematics and coding. Process reward models (PRMs) instantiate this paradigm by evaluating intermediate states. Early PRMs relied on human annotation, but recent ones emphasizes automated supervision. Specifically, Math-Shepherd \citep{wang2024math} estimates soft step-level labels via Monte Carlo sampling. OmegaPRM \citep{luoimprove} improves efficiency by localizing the earliest erroneous step using binary search. GenRM \citep{zhang2025generative} reframes verification as next-token prediction. VersaPRM \citep{zeng2025versaprm} further extends step-level supervision to multiple domains.

\smallskip
\textbf{{Atomic-Level Supervision.}}
For open-ended generation, where explicit reasoning trajectories are ill-defined, supervision shifts from global outcomes to atomic units (sentences, spans, or tokens) to better localize hallucinations, stylistic errors, and safety violations. 
At the sentence level, Fine-Grained RLHF \citep{wu2023fine} provides attribute-specific rewards, while ASPO \citep{wang2025aspo} adaptively down-weights hallucinated sentences in multi-modal reasoning. At finer resolution, token-level methods directly integrate preference signals into optimization. TDPO \citep{zengtoken} decomposes the KL regularization to the token-level for precise diversity–quality control, and TIS-DPO \citep{liu2024tis} applies importance sampling to emphasize preference-critical tokens. Complementarily, ACPO \citep{chen2025atomic} improves factuality by decomposing responses into atomic facts and scoring them via intrinsic self-consistency.

\subsection{Objective Dimensionality} 
\label{sec:dimensionality}
The final axis is objective dimensionality: whether preference signals are scalarized or preserve the multi-dimensional structure of human values.

\smallskip
\textbf{{Alignment Tax and Reward Hacking.}} 
Standard alignment pipelines often rely on scalarized objectives, encoding preference supervision as a single reward or binary label \citep{ouyang2022training, stiennon2020learning}. As alignment objectives become conflicting, this scalarization obscures context-dependent trade-offs, limiting robust generalization. One consequence is the \emph{alignment tax}, where optimizing a unified reward favors easily optimizable sub-objectives over others \citep{lin2024mitigating}, as seen in trade-offs between helpfulness and harmlessness \citep{tan2025equilibrate} or completeness and conciseness \citep{song-etal-2025-learning}. Another consequence is \emph{reward hacking}, where models exploit low-dimensional shortcuts that inflate reward scores without improving true alignment \citep{pan2024feedback}. \looseness=-1


\smallskip
\textbf{{Toward Multi-Dimensional Alignment.}}
To overcome the limitations of scalarized objectives, recent work preserves objective dimensionality by intervening at different stages of the alignment pipeline, namely response synthesis, evaluation, and optimization.

One line of work mitigates scalar collapse by restructuring the candidate behavior space at the \emph{response synthesis stage}, prior to evaluation. For example, SteerLM \citep{dong2023steerlm} and HelpSteer \citep{wang2024helpsteer, wang2024helpsteer2} condition generation on explicit control signals that specify which value dimension to emphasize, such as helpfulness and safety. This design allows scalar rewards to operate over pre-separated alignment profiles.

A complementary direction mitigates the collapse at the \emph{evaluation stage} by decomposing preference judgments into explicit criteria. Early methods such as Constitutional AI \citep{bai2022constitutional} and Self-Align \citep{sun2023principle} encode alignment objectives as normative rules guiding feedback across multiple dimensions. More recent rubric- and rationale-based evaluators, including Prometheus \citep{kim2024prometheus} and OpenRubrics \citep{liu2025openrubrics}, operationalize this idea by evaluating responses against explicit rubrics, making multi-dimensional trade-offs explicit.


At the \emph{optimization stage}, methods address conflicts by framing alignment as a multi-objective problem. Early methods include sequential alignment \citep{lou2024spo}, which optimizes objectives in separate stages, and post-hoc parameter merging \citep{jang2023personalized}, which combines models fine-tuned for different goals. 
More recent work jointly optimizes multiple objectives during training, as exemplified by MOPO \citep{agnihotri2025multi}, PAMA \citep{he2025pareto}, and MO-ODPO \citep{gupta2025robust}, enabling more principled and controllable trade-offs.

\section{Preference Instantiation Stage}
\label{sec:instanciation}


Following preference evaluation, the pipeline proceeds to preference instantiation, the translation layer between evaluation and optimization. 
This stage converts evaluated responses into structured training signals, determining how preference relations are exposed to the policy model. 
Instantiation methods are categorized by the relational complexity of feedback, ranging from \emph{(\S\ref{sec:point-wise}) point-wise} to \emph{(\S\ref{sec:pair-wise}) pair-wise} and \emph{(\S\ref{sec:group-wise}) group-wise}.

\subsection{Point-Wise Instantiation} 
\label{sec:point-wise}

Point-wise instantiation assigns an independent scalar or binary signal to each prompt –response pair $(x,y)$, avoiding explicit comparisons and enabling simple policy training aligned with many real-world feedback acquisition scenarios.

\smallskip
\textbf{{Regression-Based Supervision.}} A common approach to scalar supervision trains a separate reward model to score each prompt–response pair and optimizes the policy indirectly, as in the classical PPO family \citep{schulman2017proximal, dai2023safe, li2023remax}. Recent work improves reward fidelity to provide more reliable optimization signals and strengthen downstream alignment. For example, Critic-RM \citep{yu2025self} uses model-generated critiques to refine reward prediction, while hybrid frameworks such as HAF-RM \citep{liu2025haf} incorporate token-level supervision during reward model training. In contrast, A*PO \citep{brantley2025accelerating} skips the reward model and directly trains the policy to match offline value targets via regression.

\smallskip
\textbf{{Binary-Based Supervision.}} In contrast, binary-based supervision relies on coarse yes/no feedback that indicates whether a response is acceptable. KTO \citep{ethayarajh2024kto} relies solely on binary preference signals, applying asymmetric weighting to penalize undesirable outputs more strongly than desirable ones. Later work improves the theoretical and semantic grounding of binary supervision: BCO \citep{jung2025binary} reframes alignment from binary feedback as a classification problem, showing that the binary ones are sufficient to recover the same response ordering as DPO. RLBFF \citep{wang2025rlbffbinaryflexiblefeedback} improves binary supervision by clarifying what a yes/no label means, training the model to decide whether a response meets a specific criterion (\emph{e.g.}, correctness, clarity) instead of assigning an ambiguous binary label.

\subsection{Pair-Wise Instantiation} 
\label{sec:pair-wise}
While point-wise feedback is easy to obtain, it lacks the relational structure needed for fine-grained alignment tuning \citep{tripathi2025pairwise}. Pair-wise instantiation uses tuples $(x, y_w, y_l)$ with $y_w \succ y_l$ to model the decision boundary between preferred and dispreferred behaviors, forming the dominant paradigm in alignment tuning.

\smallskip
\textbf{{Contrastive Optimization.}} 
The standard approach in this category is DPO \citep{rafailov2023direct}, which bypasses explicit reward modeling by directly optimizing the policy with a contrastive objective comparing preferred and dispreferred responses $y_w$ and $y_l$. Despite its effectiveness, this formulation has several limitations, including overfitting and stability. IPO \citep{azar2024general} mitigates the DPO's overfitting by replacing the unbounded logit-based objective with a bounded one, while GPO \citep{tang2024generalized} unifies DPO and IPO within a generalized offline preference learning framework. In a complementary direction, sDPO \citep{kim2025sdpo} improves stability on large datasets by progressively tightening the reference policy, and MRPO \citep{le2025multi} extends the objective to multiple reference models via a stabilized virtual reference distribution.

\smallskip
\textbf{Reference-Free Optimization.} 
While effective, DPO-style methods rely on a reference model to anchor preference optimization, introducing additional overhead and potential instability. To remove this dependency, recent work derives preference signals directly from the policy's output distribution. SimPO \citep{meng2024simpo} eliminates the reference model by reinterpreting preference learning as direct optimization over the policy's own output distribution. ORPO \citep{hong2024orpo} similarly adopts a reference-free formulation, integrating preference contrast into supervised fine-tuning via an odds-ratio objective. \looseness=-1

\smallskip
\textbf{Margin-Aware Calibration.} Another limitation of standard pair-wise objectives is that they enforce relative ordering without ensuring that confidence gaps reflect preference magnitude. SLiC-HF \citep{zhao2023slic} introduces a fixed margin via a hinge loss to better align likelihood gaps with preference strength. MMPO \citep{kim2024margin} instead calibrates margins probabilistically using soft preference targets, while AlphaDPO \citep{wualphadpo} further generalizes this approach by adapting reward margins at the instance-level through implicit reference reparameterization.

\subsection{Group-Wise Instantiation}
\label{sec:group-wise}
As alignment targets shift from subjective conversation to objective reasoning, point- and pair-wise feedback become bottlenecks due to their sparsity and lack of contextual normalization. Group-wise instantiation addresses this by optimizing the policy over a candidate set $\mathbf{y}=\{y_1,\dots,y_k\}$, capturing combinatorial relationships and providing denser signals for complex preference learning.

\smallskip
\textbf{List-Wise Ranking Objectives.} 
To realize this, methods have progressed from surrogate ranking losses to direct metric optimization. RRHF \citep{yuan2023rrhf} operates as a best-of-$N$ learner using a hinge-style loss to align likelihoods with preference scores, while PRO \citep{song2024preference} refines this via iterative best-vs-rest decomposition to capture multi-level preferences. Incorporating learning-to-rank principles, LiPO \citep{liu2025lipo} applies LambdaRank-style weighting \citep{burges2006learning} to emphasize top-ranked items. Most recently, PPA \citep{zhao2025permutative} directly optimizes a differentiable NDCG objective, closing the gap between training loss and ranking metrics.

\smallskip
\textbf{Group-Relative Policy Optimization.} Beyond explicit ranking, GRPO \citep{shao2024deepseekmath} derives advantage signals from group-level statistics rather than fixed preference orders.  While this critic-free formulation reduces variance, later analyses identify estimation instability. In particular, Dr. GRPO \citep{liu2025understanding} corrects a group-normalization bias that conflates reasoning quality with response length. DAPO \citep{yu2025dapo} further mitigates entropy collapse in group-wise optimization.

\section{Key Insights and Practical Guidance}
\label{sec:insight}


Building on the unified pipeline perspective, we analyze fundamental trade-offs {(\S\ref{sec:core-trade-off})} and cross-stage interactions {(\S\ref{sec:corss_interation_main})} that govern how alignment signals are constructed and propagated, and distill a set of design principles {(\S\ref{sec:main-design-principles})} that characterize effective pipeline configurations. Finally, we translate these insights into practical, scenario-driven guidance, linking real-world constraints to concrete end-to-end pipeline design choices.

\subsection{Core Structural Trade-offs}
\label{sec:core-trade-off}

Many limitations in alignment tuning arise not from individual algorithms, but from structural trade-offs inherent to alignment data pipelines.

\smallskip
$\bullet$ \emph{Source Fidelity vs. Distribution Shift.} Offline supervision using high-quality proprietary models provides a strong, reproducible starting point but lags behind policy evolution, inducing a distribution shift as the student is evaluated on responses it cannot naturally generate. In contrast, online supervision via self-play preserves on-policy fidelity but introduces higher variance, making it prone to error amplification when the initial policy is weak. This choice shapes the exploration space and propagates to evaluation and instantiation by constraining what can be effectively judged and optimized.

\smallskip
$\bullet$ \emph{Open-Loop vs. Closed-Loop Alignment.} Many alignment pipelines operate as if in an open-loop setting, treating preference data and evaluators as static. However, alignment is inherently dynamic and policy-dependent, as evolving outputs continually reshape the data. The choice of loss influences the policy's entropy and output distribution, constraining future exploration. Ignoring this can lead to degeneracy such as reduced diversity or reward hacking, highlighting the need for adaptive mechanisms like response resampling and evaluator recalibration.

\smallskip
$\bullet$ \emph{Evaluation Granularity vs. Credit Assignment Precision.} Shifting from outcome-level to step-level evaluation improves credit assignment for complex reasoning by providing more fine-grained supervision. However, this added precision comes with substantial labeling complexity and introduces the risk of over-optimizing intermediate steps that are not strictly necessary for producing the final correct answer.

\smallskip
$\bullet$ \emph{Objective Dimensionality vs. Optimization Tax.} Collapsing multi-dimensional human values into a single scalar reward simplifies the instantiation stage but introduces an alignment tax, as important nuances across dimensions are lost. In contrast, maintaining multi-objective rubrics better preserves performance across diverse criteria, but significantly complicates the optimization landscape, making convergence slower and less stable.

\subsection{Cross-Stage Interactions}
\label{sec:corss_interation_main}



Although we categorize alignment tuning into three distinct stages---response synthesis, preference evaluation, and preference instantiation---alignment outcomes are ultimately determined by the tight coupling and interactions between them. As in Sections \ref{sec:synthesis}--\ref{sec:instanciation}, the output of one stage serves as the rigid input constraint for the next, and in iterative settings, the optimization outcome feeds back into the data generation process. This section details the three critical interaction pathways that define the efficacy of the alignment pipeline.

\subsubsection{Response Synthesis Constrains Preference Evaluation}
The quality of the preference signal is fundamentally bounded by the behavioral support generated during the response synthesis stage. No matter how sophisticated an adjudicator (in preference evaluation stage) is, it cannot extract meaningful signals from a non-informative candidate set.

\smallskip
$\bullet$ \emph{Informativeness and Selection Efficiency:} The synthesis stage must prioritize generating "informative" candidates that provide high learning value. If the synthesis strategy produces only trivial comparisons (\emph{e.g.}, a perfect response versus a nonsensical one) or ambiguous pairs with negligible quality differences, the evaluator cannot extract a strong gradient signal. Consequently, even a ground-truth evaluator becomes ineffective if the candidate set lacks the margins required to distinguish better behaviors from worse ones.

\smallskip
$\bullet$ \emph{Diversity and Contrastive Information:} Alignment tuning relies on contrast. If the synthesis strategy suffers from mode collapse or lacks exploration (as discussed in Section \ref{sec:creative_exploration}), the candidate set will consist of semantically identical responses. In this scenario, even a human expert or a strong LLM-as-a-Judge cannot assign a meaningful preference margin, leading to vanishing gradients during optimization.

\smallskip
$\bullet$ \emph{Distributional Shift and Evaluator Reliability:} The reliability of evaluation depends on how the responses are generated. If synthesized responses come from a distribution that differs substantially from the data used to calibrate the evaluator, the evaluator's confidence estimates can become misaligned. As a result, noisy or misleading scores may be injected into the optimization signal.

\subsubsection{Preference Evaluation Dictates Instantiation Capabilities}
The granularity and dimensionality of the judgments produced in the preference evaluation stage determine the upper bound of what can be modeled in the instantiation stage. A mismatch here often leads to information loss or signal distortion.

\smallskip
$\bullet$ \emph{Granularity Mismatch:} If the evaluation stage only provides outcome-level binary labels, attempting to use token-level or step-level instantiation methods (\emph{e.g.}, TIS-DPO or ACPO) is ill-posed without additional approximations. Conversely, if detailed step-level critiques are available but the instantiation method aggregates them into a single scalar reward, the dense supervision signal is compressed, losing the ability to penalize reasoning errors.

\begin{table*}[t]
\centering
\scriptsize
\renewcommand{\arraystretch}{1.4} 
\begin{tabularx}{\textwidth}{>{\raggedright\arraybackslash\hsize=0.47\hsize}X >{\raggedright\arraybackslash\hsize=1.5\hsize}X}
\toprule
{Principle} & {Key Insight} \\
\midrule
{Pipeline Defines the Optimization Signal} & 
Alignment optimizes margins induced by data, not abstract preferences. Failures often stem from weak or distorted margins in data construction rather than the loss function itself. \\ 
\hline
{Coverage Precedes Optimization} & 
Alignment is limited to sampled behaviors. Response synthesis must prioritize exploration and diversity to prevent brittle alignment on narrow supports. \\ 
\hline
{Evaluation Fidelity Sets the Upper Bound} & 
Evaluator reliability (calibration, consistency) determines alignment quality. Improving judge fidelity often yields larger gains than modifying downstream objectives. \\ 
\hline
{Granularity Enables Credit Assignment} & 
Outcome-level supervision is insufficient for complex reasoning. Step- or token-level granularity is required to localize errors and reduce spurious correlations. \\ 
\hline
{Preserve Preference Structure} & 
Scalar rewards obscure relational trade-offs. Pairwise, group-wise, and list-wise formulations better preserve the multi-dimensional structure of human preferences. \\ 
\hline
{Alignment is a Closed-Loop Design Problem\!\!\!\!\!} & 
Data is policy-dependent. Effective pipelines must be adaptive, iteratively reshaping the preference landscape as the policy improves. \\ 
\bottomrule
\end{tabularx}
\vspace*{-0.25cm}
\caption{Design principles for alignment data pipelines. We distill six key principles governing response synthesis, preference evaluation, and instantiation.}
\vspace*{-0.2cm}
\label{tab:alignment_principles}
\end{table*}

\smallskip
$\bullet$ \emph{Dimensionality and Alignment Tax:} When evaluation captures multi-dimensional trade-offs (\emph{e.g.}, helpfulness vs. safety), but the instantiation layer collapses these into a single scalarized value, it forces the policy to optimize a fixed trade-off. This often results in the "alignment tax," where improvements in one dimension inadvertently degrade performance in another due to the inability of the scalar loss to represent the Pareto frontier.

\subsubsection{Preference Instantiation Reshapes Future Response Synthesis}
In iterative or online alignment settings, the pipeline operates as a closed loop. The choice of loss function in the preference instantiation stage directly impacts the policy model's entropy, which in turn defines the search space for the next round of response synthesis.

\smallskip
$\bullet$ \emph{Mode Collapse and Exploration:} Contrastive objectives like DPO effectively increase the likelihood of preferred responses but can rapidly reduce the entropy of the policy. If the instantiation stage enforces margins too aggressively, the updated policy model may lose the diversity required for effective exploration in the subsequent synthesis step. This creates a degenerate cycle where the model stops generating novel negatives, starving the pipeline of the informative data needed for further improvement.

\smallskip
$\bullet$ \emph{Reward Hacking Dynamics:} If the instantiation mechanism has exploitable shortcuts, the policy will learn to maximize the score without improving true quality. Over time, it produces responses that look good according to the reward but are actually poor. This makes evaluation increasingly difficult and often requires the evaluator to be updated or recalibrated.

\subsection{Pipeline Design Principles}
\label{sec:main-design-principles}

The trade-offs and interactions motivate design principles for alignment data pipelines. Rather than prescribing specific algorithms, these principles characterize how response synthesis, preference evaluation, and preference instantiation should be jointly designed to manage these trade-offs. Based on our analysis of existing methods, we distill six principles to capture effective design patterns across response synthesis, response evaluation, and response instantiation. We summarize the principles and their core insights in Table \ref{tab:alignment_principles}, which serves as a unifying reference for prior work.



Beyond conceptual principles above, translating these insights into practice requires addressing real-world constraints such as cost, latency, hardware limits, and labeling budgets. To make this survey a practical, we map alignment methods to design scenarios shaped by resource limitations and system-level trade-offs. Therefore, we provide stage-wise decision frameworks linking these practical constraints to specific alignment design choices.

\smallskip
$\bullet$ \emph{Response Synthesis Stage (Constraints on Exploration).}  The primary function of this stage is to define the behavioral support of the policy. The design choice is governed by Data Availability (Does the data exist?) and Inference Budget (Can we afford to generate it?). Table \ref{tab:synthesis_constraints} outlines how different constraints lead to distinct synthesis strategies, each with inherent trade-offs between efficiency, distribution fidelity, and exploration.

\begin{table*}[t!]
\centering
\scriptsize
\begin{tabular}{@{} >{\raggedright\arraybackslash}p{0.22\textwidth} >{\raggedright\arraybackslash}p{0.20\textwidth} >{\raggedright\arraybackslash}p{0.37\textwidth} >{\raggedright\arraybackslash}p{0.12\textwidth} @{}}
\toprule
{Constraint / Scenario} & {Recommended Strategy} & {Design Rationale \& Trade-offs} & {Relevant Methods} \\
\midrule
{Low Inference Budget}\newline (Cannot afford online generation) & Offline Synthesis\newline (Policy-Aware Reweighting) & {Rationale:} Uses existing static datasets to save compute.\newline {Trade-off:} Risk of distribution shift (off-policy); requires reweighting to correct bias. & WPO, WRPO, TIS-DPO \\
\midrule
{New Domain / Cold Start}\newline (No existing preference data) & Online Synthesis\newline (Iterative Self-Play) & {Rationale:} Essential when no in-domain data exists; the model generates its own training data to bridge distribution shift.\newline {Trade-off:} High training latency due to iterative generation steps. & SPIN, SPPO, Iterative DPO, SELM \\
\midrule
{Mode Collapse / Repetitive Outputs}\newline (Lack of diversity) & Creative Exploration\newline (Verbalized / Diversity Sampling) & {Rationale:} Standard sampling narrows the search space; explicit exploration preserves behavioral diversity. & Verbalized Sampling, Spectrum Tuning, DivPO, CRPO \\
\bottomrule
\end{tabular}
\vspace*{-0.25cm}
\caption{Response synthesis strategies under real-world constraints.}
\vspace*{-0.2cm}
\label{tab:synthesis_constraints}
\end{table*}

\begin{table*}[t!]
\centering
\scriptsize
\begin{tabular}{@{} >{\raggedright\arraybackslash}p{0.22\textwidth} >{\raggedright\arraybackslash}p{0.20\textwidth} >{\raggedright\arraybackslash}p{0.37\textwidth} >{\raggedright\arraybackslash}p{0.12\textwidth} @{}}
\toprule
{Constraint / Scenario} & {Recommended Adjudicator \& Granularity} & {Design Rationale \& Trade-offs} & {Relevant Methods} \\
\midrule
{Low Labeling Budget}\newline (Cannot afford Human/API judges) & Weak Model\newline (Outcome-Level) & {Rationale:} Cost-effective for simple, general tasks.\newline {Trade-off:} Noisy signals; often requires aggregation or collective intelligence to be reliable. & Meta-Rater, RLAIF \\
\midrule
{High Task Complexity}\newline (Multi-step reasoning tasks) & Verifiers\newline (Step-Level) & {Rationale:} Outcome-level labels fail to localize logic errors; step-level signals are required for precise credit assignment.\newline {Trade-off:} Requires domain-specific verifiers or ground-truth checkers. & Math-Shepherd, OmegaPRM, VersaPRM, GenRM \\
\midrule
{Conflicting Objectives}\newline (e.g., Safety vs. Utility) & Multi-Criteria Rubric\newline (Multi-Dimensional) & {Rationale:} A single scalar score obscures trade-offs; explicit dimension separation is required to manage competing goals. & Prometheus, HelpSteer, OpenRubrics \\
\bottomrule
\end{tabular}
\vspace*{-0.25cm}
\caption{Preference evaluation strategies under real-world constraints.}
\vspace*{-0.2cm}
\label{tab:evaluation_constraints}
\end{table*}

\smallskip
$\bullet$ \emph{Preference Evaluation Stage (Constraints on Fidelity).} The primary function of this stage is to assign preference judgments via an adjudicator. The design choice is governed by Task Complexity (How hard is it to evaluate?), Labeling Budget (Can we afford high-quality judges?), and Objective Dimensionality (Are there conflicting goals like safety vs. helpfulness?).  Table \ref{tab:evaluation_constraints} organizes these constraints into corresponding choices of adjudicator and granularity, highlighting the trade-offs between cost and evaluation precision.

\smallskip
$\bullet$ \emph{Preference Instantiation Stage (Constraints on Optimization).} The primary function of this stage is to convert evaluated responses into structured training signals. The design choice is governed by Hardware Constraints (VRAM), Task Variance (Instabilities in optimization), and Objective Dimensionality (Conflicting goals). Table \ref{tab:instantiation_constraints} organizes these constraints into corresponding feedback instantiation strategies, highlighting the trade-offs between computational efficiency, optimization stability, and signal expressiveness.

Additionally, Appendix \ref{sec:guidance} presents scenario-driven pipeline configurations that jointly integrate response synthesis, preference evaluation, and instantiation, enabling end-to-end design under realistic deployment constraints.

\section{Future Research Directions}
\label{sec:future_challenge}

Despite substantial progress in alignment tuning, several fundamental challenges remain unresolved, particularly as alignment settings become more dynamic, complex, and deployment-driven. To address these limitations, we highlight four key directions for future research that aim to move beyond current pipeline paradigms and enable more adaptive, robust, and scalable alignment frameworks.

$\bullet$ \emph{{Prompt-Level Alignment.}}
Alignment must move beyond global preference models toward prompt-level alignment. Different prompts inherently require different evaluation criteria such as safety creativity or precision. However current pipelines apply a shared preference function across heterogeneous prompts. A central challenge is inferring and enforcing prompt specific alignment criteria without explicit human specification.

$\bullet$ \emph{{Alignment for Agentic Systems.}}
As agentic systems become dominant alignment must move beyond individual decisions to agentic loops. In long horizon workflows failures arise from interactions among planning execution feedback and memory rather than single actions. Existing methods supervise local outputs while leaving global behavior unconstrained. A key challenge is defining trajectory-level alignment objectives that enable agents to revise or abandon plans over time.

\begin{table*}[t!]
\centering
\scriptsize
\begin{tabular}{@{} >{\raggedright\arraybackslash}p{0.22\textwidth} >{\raggedright\arraybackslash}p{0.20\textwidth} >{\raggedright\arraybackslash}p{0.37\textwidth} >{\raggedright\arraybackslash}p{0.12\textwidth} @{}}
\toprule
{Constraint / Scenario} & {Feedback Instantiation} & {Design Rationale \& Trade-offs} & {Relevant Methods} \\
\midrule
{Strict Hardware Constraints}\newline (Low VRAM) & Pair-wise\newline (Reference-Free) & {Rationale:} Removes the memory bottleneck and computational overhead of loading a frozen reference model. & DPO, SimPO, ORPO \\
\midrule
{High Task Variance} & Group-wise & {Rationale:} Normalizing rewards across a group of sampled candidates ($N>2$) stabilizes gradients and reduces variance compared to pairwise contrasts. & GRPO, LiPO, Dr. GRPO, DAPO \\
\midrule
{Conflicting Objectives} & Multi-Objective & {Rationale:} Optimizes multiple objectives jointly to model the Pareto frontier directly, rather than collapsing multi-dimensional trade-offs into a single scalar reward. & MOPO, PAMA, MO-DPO \\
\bottomrule
\end{tabular}
\vspace*{-0.25cm}
\caption{Preference instantiation strategies under real-world constraints.}
\vspace*{-0.2cm}
\label{tab:instantiation_constraints}
\end{table*}

$\bullet$ \emph{{Alignment under Evolving Objectives.}}
Future alignment must account for objectives that evolve over time rather than remain fixed. In interactive and agentic settings user intent task goals and acceptable behavior change during deployment. Current pipelines assume static training time preferences. A key challenge is adapting alignment to evolving objectives without destabilizing learned behavior or inducing value drift.

$\bullet$ \emph{{Multi-Modal Alignment.}} As LLMs evolve into large multi-modal models, alignment must extend beyond text. Modalities such as video introduce tightly coupled temporal, spatial, and visual structures that are difficult to control with coarse supervision. A key challenge is designing cross-modal alignment pipelines that translate high-level feedback into fine-grained updates over multi-modal representations.

\section{Conclusion}
\label{sec:conclusion}

This survey reframes alignment tuning as a data-centric pipeline design problem. By decomposing alignment into response synthesis, preference evaluation, and preference instantiation, we show how pipeline design governs alignment behavior. Our analysis reveals recurring trade-offs and interactions that guide future alignment pipelines.

\bibliographystyle{assets/plainnat}
\bibliography{colm2026_conference}

\appendix
\clearpage

\begin{table}[htbp]
\centering
\label{tab:methodology-comparison}
\resizebox{\textwidth}{!}{
\begin{tabular}{c|ccc|cc|c}
\toprule
& \multicolumn{3}{c}{Response Synthesis} &  \multicolumn{2}{|c|}{Preference Evaluation} & Preference \\
{Methodology} & {Response Source} & {Selection Strategy} & {Creative Exploration} & {Judgement Granularity} & {Object Dimensionality} & {Instantiation} \\
\midrule
WPO & Offline & Random & None & Outcome-level & Single-dim & Point-wise \\
WRPO & Offline & Random & None & Outcome-level & Single-dim & Point-wise \\
TIS-DPO & Offline & Random & None & Atomic-level & Single-dim & Pair-wise \\
Iterative DPO & Online & Random & None & Outcome-level & Single-dim & Pair-wise \\
RS-DPO & Online & Margin-based & None & Outcome-level & Single-dim & Pair-wise \\
SPIN & Online & Random & None & Outcome-level & Single-dim & Pair-wise \\
SPPO & Online & Random & None & Outcome-level & Single-dim & Group-wise \\
GRPO & Online & Random & None & Outcome-level & Single-dim & Group-wise \\
SELM & Online & Random & None & Outcome-level & Single-dim & Pair-wise \\
BeeS & Offline & Margin-based & None & Outcome-level & Single-dim & Pair-wise \\
MMPO & Offline & Margin-based & None & Outcome-level & Single-dim & Pair-wise \\
AlphaDPO & Offline & Margin-based & None & Outcome-level & Single-dim & Pair-wise \\
APL & Online & Uncertainty-based & None & Outcome-level & Single-dim & Pair-wise \\
MAPLE & Online & Uncertainty-based & None & Outcome-level & Single-dim & Pair-wise \\
IUPO & Online & Uncertainty-based & None & Step-level & Single-dim & Pair-wise \\
UPO & Online & Uncertainty-based & None & Outcome-level & Single-dim & Pair-wise \\
UDASA & Online & Uncertainty-based & None & Outcome-level & Multi-dim & Pair-wise \\
Verbalized Sampling & Online & Random & Verbalized Sampling & Outcome-level & Single-dim & Pair-wise \\
Spectrum Tuning & Offline & Random & Explicit Spanning & Outcome-level & Single-dim & Point-wise \\
DivPO & Online & Margin-based & Diversity Penalty & Outcome-level & Single-dim & Pair-wise \\
CRPO & Online & Random & Multi-signal & Outcome-level & Multi-dim & Pair-wise \\
Meta-Rater & Offline & Random & None & Outcome-level & Multi-dim & Point-wise \\
OmegaPRM & Offline & Random & None & Step-level & Single-dim & Point-wise \\
GenRM & Offline & Random & None & Step-level & Single-dim & Point-wise \\
VersaPRM & Offline & Random & None & Step-level & Single-dim & Point-wise \\
Math-Shepherd & Offline & Random & None & Step-level & Single-dim & Point-wise \\
ASPO & Offline & Random & None & Atomic-level & Single-dim & Pair-wise \\
TDPO & Offline & Random & None & Atomic-level & Single-dim & Pair-wise \\
SteerLM & Offline & Random & None & Outcome-level & Multi-dim & Point-wise \\
HelpSteer & Offline & Random & None & Outcome-level & Multi-dim & Point-wise \\
Prometheus & Offline & Random & None & Outcome-level & Multi-dim & Point-wise \\
PPO & Online & Random & None & Outcome-level & Single-dim & Point-wise \\
KTO & Offline & Random & None & Outcome-level & Single-dim & Point-wise \\
DPO & Offline & Random & None & Outcome-level & Single-dim & Pair-wise \\
SimPO & Offline & Random & None & Outcome-level & Single-dim & Pair-wise \\
ORPO & Offline & Random & None & Outcome-level & Single-dim & Pair-wise \\
IPO & Offline & Random & None & Outcome-level & Single-dim & Pair-wise \\
RRHF & Offline & Random & None & Outcome-level & Single-dim & Group-wise \\
PRO & Offline & Random & None & Outcome-level & Single-dim & Group-wise \\
LiPO & Offline & Random & None & Outcome-level & Single-dim & Group-wise \\
PPA & Offline & Random & None & Outcome-level & Single-dim & Group-wise \\
\bottomrule
\end{tabular}
}
\vspace*{-0.2cm}
\caption{Categorization of alignment tuning methods across three data pipeline stages.}
\vspace*{-0.2cm}
\label{tab:categorization}
\end{table}

\section{Alignment Tuning Algorithms}
\label{sec:algo_detail}

\emph{Proximal Policy Optimization (PPO)} is a representative approach for explicitly maximizing a learned reward under the KL-regularized objective in Eq.~(\ref{eq:alignment_tuning}) \citep{schulman2017proximal}. 
It proceeds in two stages.
First, a reward model $r_\phi(x,y)$ is trained from pairwise comparisons to approximate oracle preferences $r^{*}(x,y)$.
Second, the policy $\pi_\theta$ is optimized with reinforcement learning to maximize the learned reward while staying close to a reference policy $\pi_{\mathrm{ref}}$:
\begin{equation*}
\max_{\pi_\theta}\;
\mathbb{E}_{x,y}\!\left[
r_\phi(x,y) - \beta \log \frac{\pi_\theta(y\mid x)}{\pi_{\mathrm{ref}}(y\mid x)}
\right].
\end{equation*}
In practice, PPO performs stable policy updates using a clipped surrogate objective,
\begin{equation*}
\small
\begin{aligned}
&\mathcal{L}_{\mathrm{PPO}}(\theta)
=
\mathbb{E}_{x,y}\!\Big[
\min\!\Big(
\rho_\theta(x,y)\,\hat{A}(x,y), \\
&\qquad\qquad
\mathrm{clip}(\rho_\theta(x,y),1-\epsilon,1+\epsilon)\,\hat{A}(x,y)
\Big)
\Big].
\end{aligned}
\end{equation*}
where $\rho_\theta(x,y)=\frac{\pi_\theta(y\mid x)}{\pi_{\theta_{\mathrm{old}}}(y\mid x)}$ is the likelihood ratio and $\hat{A}(x,y)$ is an advantage estimate constructed from $r_\phi(x,y)$ (typically with a value baseline).
The clipping term constrains the effective step size, preventing large deviations from the reference distribution and improving training stability.

\emph{Direct Preference Optimization (DPO)} bypasses explicit reward modeling by directly optimizing the KL-regularized objective in Eq.~(\ref{eq:alignment_tuning}) using preference pairs \citep{rafailov2023direct}.
Given a dataset of comparisons $(x,y_w,y_l)$ where $y_w \succ y_l$, DPO yields a classification-style objective that increases the relative likelihood of preferred responses under $\pi_\theta$ compared to $\pi_{\mathrm{ref}}$:
\begin{equation}
\small
\begin{aligned}
\mathcal{L}_{\mathrm{DPO}}(\theta)
&=
-\mathbb{E}_{(x,y_w,y_l)}
\Big[
\log \sigma\!\Big(
\beta \Big(
\log \tfrac{\pi_\theta(y_w\mid x)}{\pi_{\mathrm{ref}}(y_w\mid x)}
\\
&\qquad\qquad\qquad
-
\log \tfrac{\pi_\theta(y_l\mid x)}{\pi_{\mathrm{ref}}(y_l\mid x)}
\Big)
\Big)
\Big].
\end{aligned}
\end{equation}
where $\sigma(\cdot)$ is the logistic sigmoid.
This objective can be interpreted as implicitly fitting a reward via the log-likelihood ratio
$\log \frac{\pi_\theta(y\mid x)}{\pi_{\mathrm{ref}}(y\mid x)}$,
thereby achieving preference optimization without training a separate reward model.

\emph{Group Relative Policy Optimization (GRPO)} shifts from pairwise comparisons to group-wise optimization by normalizing candidates sampled from the current policy \citep{shao2024deepseekmath}.
For each prompt $x$, GRPO samples a set of responses $\{y_i\}_{i=1}^{k} \sim \pi_\theta(\cdot\mid x)$ and computes a group-relative baseline (\emph{e.g.}, the mean score):
\begin{equation*}
\small
\bar{r}(x)=\frac{1}{k}\sum_{i=1}^{k} r(x,y_i),
\end{equation*}
where $r(x,y)$ is a scoring function (often derived from verification) used to rank candidates.
The policy is then updated to increase the probability of responses that outperform the group baseline:
\begin{equation*}
\small
\max_{\pi_\theta}\;
\mathbb{E}_{x}\,\mathbb{E}_{y\sim \pi_\theta(\cdot\mid x)}
\left[
\bigl(r(x,y)-\bar{r}(x)\bigr)\,
\log \pi_\theta(y\mid x)
\right].
\end{equation*}
By using group-relative normalization, GRPO reduces variance and avoids the need for an explicit critic, value networks.

\emph{Other Algorithms.}
Building on these foundational approaches, the alignment literature has expanded into several related families that explore alternative design choices and practical trade-offs.

Methods in the PPO lineage continue to emphasize training stability and sample efficiency under explicit reward maximization, with extensions such as Safe-RLHF \citep{dai2023safe} incorporating safety-aware constraints and ReMax \citep{li2023remax} removing the need for an explicit critic.

The DPO family investigates simplifications of preference optimization under different supervision assumptions: SimPO \citep{meng2024simpo} eliminates the reference policy to reduce computational overhead, while KTO \citep{ethayarajh2024kto} relaxes the dependence on paired preference data by leveraging binary feedback.

Finally, the GRPO family generalizes preference optimization to group-wise settings. DAPO \citep{yu2025dapo} mitigates entropy collapse through decoupled clipping, whereas Dr. GRPO \citep{liu2025understanding} improves computational efficiency by removing biased normalization terms.

\begin{table*}[t]
\centering
\small
\renewcommand{\arraystretch}{1.2}
\resizebox{\textwidth}{!}{%
\begin{tabular}{p{3.5cm} p{3.0cm} p{3.0cm} p{3.0cm} p{4.5cm}}
\toprule
\textbf{Scenario / Constraints} & \textbf{Response Synthesis} & \textbf{Preference Evaluation} & \textbf{Preference Instantiation} & \textbf{Rationale \& Methods} \\
\midrule
\textbf{Resource-Constrained} \newline \textit{(Low Budget, General Chat)} & 
\textbf{Offline} \newline Use existing off-policy datasets \newline(e.g., UltraFeedback). & 
\textbf{Heuristic / None} \newline Rely on pre-annotated labels; \newline avoid re-labeling. & 
\textbf{Pair-wise (Ref-free)} \newline No separate Reward \newline Model to save VRAM. & 
\textbf{Why:} Maximizes memory \newline efficiency by removing reference \newline model loading. \newline \textbf{Methods:} SimPO, ORPO. \\
\midrule
\textbf{Complex Reasoning} \newline \textit{(Math, Coding, Logic)} & 
\textbf{Online (Sample)} \newline Generate multiple rollouts per prompt \newline($N>1$). & 
\textbf{Step-level / Verifiable} \newline Use compilers or \newline ground-truth checkers. & 
\textbf{Group-wise} \newline Normalize rewards \newline within a group. & 
\textbf{Why:} Group-relative signals reduce variance; step-level supervision localizes logic errors. \newline \textbf{Methods:} GRPO, Process Rewards. \\
\midrule
\textbf{Open-Ended Creativity} \newline \textit{(Storytelling, Roleplay)} & 
\textbf{Offline + Creativity} \newline High-temp sampling or diversity constraints. & 
\textbf{LLM-as-a-Judge} \newline Use strong models \newline(e.g., GPT-4) to rank. & 
\textbf{List-wise / Pair-wise} \newline Capture nuances \newline beyond binary good/bad.\!\!\! & 
\textbf{Why:} List-wise ranking preserves diversity better than scalar regression; avoids mode collapse. \newline \textbf{Methods:} LiPO, DivPO. \\
\midrule
\textbf{Strict Compliance} \newline \textit{(Safety, Harmlessness)} & 
\textbf{Offline (Targeted)} \newline Curate red-teaming \newline prompts. & 
\textbf{Multi-Criteria} \newline Separate signals for \newline Helpfulness \& Safety.\!\!\! & 
\textbf{Multi-Objective} \newline Optimize utility \newline s.t. safety constraints. & 
\textbf{Why:} Scalarizing conflicting objectives leads to jailbreaks; constrained optimization enforces boundaries. \newline \textbf{Methods:} Safe-RLHF, MOPO. \\
\midrule
\textbf{Data-Poor / Cold Start} \newline \textit{(New Domain Adaptation)} & 
\textbf{Online (Iterative)} \newline Self-play from current \newline policy. & 
\textbf{LLM-as-a-Judge} \newline Use generic strong LLM\!\!\newline to label on-policy data. & 
\textbf{Iterative Pair-wise} \newline Re-train on batches of self-generated data. & 
\textbf{Why:} Bridges distribution shift; model learns from its own winning responses. \newline \textbf{Methods:} SPPO, Iterative DPO. \\
\bottomrule
\end{tabular}%
}
\vspace*{-0.25cm}
\caption{Recommended alignment pipeline configurations based on resource and task constraints. We map common deployment scenarios to the most effective combination of pipeline components (response synthesis, evaluation, and instantiation) as discussed in the main text.}
\vspace*{-0.2cm}
\label{tab:pipeline_guidelines_2}
\end{table*}

\section{Practical Guidelines for Alignment Pipeline Design}
\label{sec:guidance}

While the main body of this survey systematizes the design space of alignment tuning along response synthesis, preference evaluation, and instantiation, practitioners often face a more immediate question: "which pipeline configuration should be selected under concrete operational constraints?"
To bridge this gap between conceptual taxonomy and real-world deployment, we provide practical guidelines for alignment pipeline design based on three governing factors: data availability, task complexity, and computational budget.

Table \ref{tab:pipeline_guidelines_2} summarizes recommended pipeline configurations for common deployment scenarios. This section explicates the rationale behind each recommendation, clarifying how different pipeline components should be combined in practice.

\subsection{Decision Factors}

Alignment pipeline design is primarily constrained by three factors: data availability, task complexity, and computational budget.
These factors directly shape feasible choices across response synthesis, preference evaluation, and instantiation.

\paragraph{Data Availability.}
The availability and reliability of supervision signals constrain pipeline design. When large-scale, high-quality preference data exists, offline alignment using pre-collected datasets is often sufficient and cost-effective. In contrast, data-poor or cold-start settings require on-policy data generation and iterative self-improvement, as the model must construct its own supervision signals to overcome distribution shift.

\paragraph{Task Complexity.}
Task structure determines the appropriate evaluation granularity.
Tasks with explicit correctness criteria, such as mathematics, coding, or formal logic, benefit from trajectory-level or verifiable supervision that localizes errors within intermediate steps.
Conversely, open-ended tasks lack objective ground-truth and require comparative judgments that preserve diversity and relative quality rather than absolute correctness.

\paragraph{Computational Budget.}
Compute and memory constraints strongly influence feasible pipeline components. Under limited budgets, pipelines that avoid auxiliary models and online sampling are preferred. When sufficient compute is available, richer evaluation signals and multi-rollout sampling can significantly improve alignment stability and sample efficiency.

\subsection{Scenario-Specific Design Rationale}

We next explain the design rationale behind each scenario listed in Table \ref{tab:alignment_principles}, clarifying why particular combinations of synthesis, evaluation, and instantiation are well suited to each setting.

\paragraph{Resource-Constrained Setup.}
In low-budget general chat settings, offline response synthesis using existing off-policy datasets minimizes data collection cost while maintaining reasonable coverage of user behaviors. Heuristic or pre-annotated preference evaluation avoids repeated labeling, while reference-free pair-wise instantiation removes the need for a separate reward or reference model. This configuration maximizes memory efficiency and supports stable alignment under strict constraints.

\paragraph{Complex Reasoning Tasks.}
For tasks involving structured reasoning, multiple rollouts per prompt are necessary to expose alternative solution trajectories. Trajectory-level or verifier-based evaluation, such as compilers or ground-truth checkers, provides precise supervision by identifying where reasoning succeeds or fails. Group-wise instantiation normalizes rewards within a cohort of candidates, reducing variance and stabilizing optimization as reasoning difficulty varies across samples.

\paragraph{Open-Ended Creative Generation.}
Creative tasks such as storytelling or roleplay lack objective correctness and are sensitive to mode collapse. High-temperature or diversity-constrained synthesis encourages stylistic variation, while LLM-as-a-Judge evaluation provides relative quality judgments beyond binary acceptability. List-wise or pair-wise instantiation preserves nuanced preferences among candidates, maintaining diversity more effectively than scalar objectives.

\paragraph{Strict Compliance and Safety Alignment.}
Safety-critical deployment requires disentangling competing objectives such as helpfulness and harmlessness. Targeted offline synthesis using red-teaming prompts focuses supervision on failure modes. Multi-criteria evaluation separates safety and utility signals, while multi-objective instantiation enforces explicit safety boundaries that scalar objectives often fail to maintain.

\paragraph{Data-Poor and Cold-Start Adaptation.}
In new domains where external supervision is scarce, iterative online alignment becomes essential. Self-play from the current policy generates on-policy data, which is labeled using a general-purpose strong LLM as a judge. Iterative pair-wise instantiation allows the model to repeatedly learn from its own highest-quality outputs, gradually bridging distribution shift and accelerating adaptation.

\subsection{Takeaway}
Across scenarios, the key insight is that effective alignment is not determined by a single algorithmic choice, but by the {coherent co-design} of synthesis, evaluation, and instantiation under practical constraints. Table \ref{tab:alignment_principles} should therefore be read not as a prescriptive checklist, but as a set of principled templates that can be adapted to the specific operational realities of a given deployment. While these principles are not universally optimal in all settings, they provide a coarse yet informative abstraction of recurring design trade-offs observed across alignment pipelines.

\end{document}